%% file: main.tex
\documentclass[runningheads]{llncs}

\usepackage{eccv}

\usepackage{eccvabbrv}

\usepackage{graphicx}
\usepackage{booktabs}
\usepackage{siunitx}
\usepackage{url}
\usepackage[accsupp]{axessibility}  
\usepackage{hyperref}
\newcommand{\tai}{\textcolor{black}}
\newcommand{\name}{OmniSch}

\usepackage{booktabs}
\usepackage{multirow}
\usepackage{amssymb}
\usepackage{xcolor}
\usepackage{makecell}
\usepackage{algorithm}
\usepackage{algpseudocode}

\usepackage{array}
\usepackage{booktabs}
\usepackage{tabularx}
\usepackage{pifont}
\usepackage[table]{xcolor}

\newcommand{\cmark}{\textcolor{green!60!black}{\ding{51}}}
\newcommand{\xmark}{\textcolor{red}{\ding{55}}}

\definecolor{rowblue}{RGB}{220,230,240}
\usepackage[table]{xcolor}

\definecolor{bestblue}{RGB}{0,70,180}
\definecolor{secondgreen}{RGB}{0,150,0}

\newcommand{\pos}[1]{\textcolor{green!60!black}{+\num{#1}\%}}
\newcommand{\negv}[1]{\textcolor{red}{-\num{#1}\%}}

\newcommand{\best}[1]{\textbf{#1}}
\newcommand{\second}[1]{\underline{#1}}

\usepackage[table]{xcolor}
\newcommand{\graycell}{\cellcolor{gray!20}}

\usepackage{orcidlink}

\title{OmniSch: A Multimodal PCB Schematic Benchmark For Structured Diagram Visual Reasoning} 

\titlerunning{OmniSch: A Multimodal PCB Schematic Benchmark}

\author{
Taiting Lu\inst{1}\textsuperscript{*} \and
Kaiyuan Lin\inst{1}\textsuperscript{*} \and
Yuxin Tian\inst{2} \and
Mingjia Wang\inst{4} \and
Yubo Wang\inst{1} \and
Muchuan Wang\inst{2} \and
Sharique Khatri\inst{1} \and
Akshit Kartik\inst{1} \and
Yixi Wang\inst{1} \and
Amey Santosh Rane\inst{3} \and
Yida Wang\inst{4} \and
Sung-Liang Chen\inst{4} \and
Yifan Yang\inst{5}\textsuperscript{†} \and
Yi-Chao Chen\inst{4} \and
Yincheng Jin\inst{3} \and
Mahanth Gowda\inst{1}\textsuperscript{†}
}

\authorrunning{Lu and Lin et al.}

\institute{
Pennsylvania State University, USA 
\email{\{txl5518, kjl6302, axk6143, sqk6377, ymw5456, yubow, mahanth.gowda\}@psu.edu}
\and
Independent Researcher 
\email{\{chelseatiann@gmail.com, starboymc329666@outlook.com\}}
\and
Binghamton University, USA 
\email{\{yjin5, arane4\}@binghamton.edu}
\and
Shanghai Jiao Tong University, China 
\email{\{mingjiawang, yidawang, sungliang.chen, yichao\}@sjtu.edu.cn}
\and
Microsoft Research
\email{yifanyang@microsoft.com}
}

\begin{document}

\maketitle

\begingroup
\renewcommand\thefootnote{}
\footnotetext{* Both authors contributed equally.}
\footnotetext{† Corresponding author.}
\footnotetext{Project Page: \url{https://omnisch.com/}}
\endgroup

\begin{abstract}
    
    Recent large multimodal models (LMMs) have made rapid progress in visual grounding, document understanding, and diagram reasoning tasks. 
    However, their ability to convert Printed Circuit Board (PCB) schematic diagrams into machine-readable spatially weighted netlist graphs, jointly capturing component attributes, connectivity, and geometry, remains largely underexplored, despite such graph representations are the backbone of practical electronic design automation (EDA) workflows.
    To bridge this gap, we introduce \textbf{\name}, the first comprehensive benchmark designed to assess LMMs on schematic understanding and spatial netlist graph construction. 
    {\name} contains 1,854 real-world schematic diagrams and includes four tasks: \textbf{(1) visual grounding} for schematic entities, with 109,9\textit{K} grounded instances aligning 423.4\textit{K} diagram semantic labels to their visual regions; \textbf{(2) diagram-to-graph reasoning}, understanding topological relationship among diagram elements;
    \textbf{(3) geometric reasoning}, constructing layout-dependent weights for each connection; and \textbf{(4) tool-augmented agentic reasoning} for visual search, invoking external tools to accomplish (1)-(3).
    Our results reveal substantial gaps of current LMMs in interpreting schematic engineering artifacts, including unreliable fine-grained grounding, brittle layout-to-graph parsing, inconsistent global connectivity reasoning and inefficient visual exploration. 


\end{abstract}

\input{intro}

\input{related_work}

\input{benchmark}

\input{experiment}

\section{Conclusion}
In this work, we introduce {\name}, a large-scale real-world schematic benchmark with rich symbol, pin, text, and net graph annotations, along with structured evaluation protocols for assessing netlist graph generation by LMMs.
Leveraging this benchmark, we conduct extensive experiments on representative LMMs across diverse evaluation settings, from single-shot prompting to tool-augmented agentic pipelines.
Based on the results, we identify essential factors that affect their performance.
We observed a tool-augmented agentic framework could enable LMMs to perform more active visual search through external perception tools, improving the efficiency and accuracy of schematic-to-netlist generation.
We hope {\name} benchmark could support further research on improving LMMs' structured diagram understanding capability.

\newpage

%

%
\bibliographystyle{splncs04}
\bibliography{main}

\clearpage

\end{document}

%% file: intro.tex
\section{Introduction}
\label{sec:intro}
Printed Circuit Board (PCB) schematic diagrams are essential design representations for electronic design automation (EDA) systems, spanning from daily smartphones to industrial automotive platforms. 
However, most real-world schematic designs are proprietary and rarely publicly available, and even when accessible they require labor-intensive expert annotation, leaving most open design knowledge confined to textbooks, research papers, and patents, where circuits are presented as human-readable diagrams rather than machine-parsable netlists.
To enable downstream EDA workflows, schematic diagrams must be converted into machine-readable netlists that preserve component connectivity and parameters for analysis, simulation, and layout synthesis.
Recently, several works \cite{huang2025netlistify,bhandari2024masala,xu2025image2net} have employed YOLO models \cite{cheng2024yoloworld} to automate this conversion from textbook schematics to netlists and released public datasets. 
As shown in Table \ref{tab:pcb_benchmark}, existing datasets fall short of real-world complexity, covering only a limited number of entity types and samples compared to the thousands of component types in practical designs.


\begin{figure}[t]
    \centering
    \includegraphics[width=\linewidth]{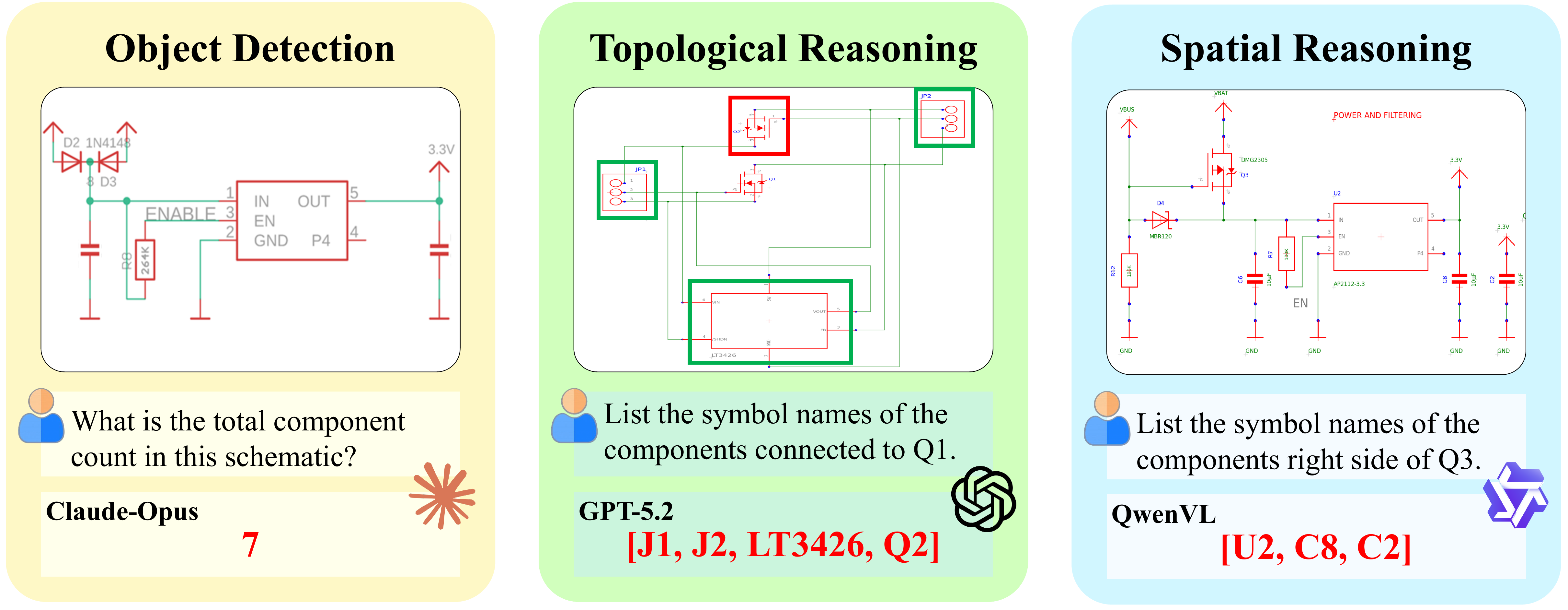}
    \caption{\textbf{Large multimodal models fail to reliably perform core visual understanding tasks on structured schematic diagrams.} Errors in component detection, topological connectivity reasoning, and spatial layout interpretation highlight persistent limitations in handling the compositional and relational complexity inherent in schematic diagrams.}
    \label{fig:lmm_test}
\end{figure}

In recent years, large multimodal models (LMMs), such  GPT-5 \cite{openai_gpt5} and Gemini 2.5 \cite{google_gemini_25_flash}, have demonstrated strong open-ended capabilities in geometric reasoning, multimodal grounding, and graph-based understanding.
These advances suggest new opportunities for applying LMMs to engineering workflows, particularly for structured visual artifacts such as PCB schematic diagrams and generating their corresponding netlists graph. 
However, schematic-to-netlist graph generation remains largely underexplored and unbenchmarked, raising a key question: \textit{could current LMMs reliably recover component attributes, find-grained level connectivity, and geometry-aware edges from real-world, densely cluttered schematics?}


To answer the question above, we conducted preliminary tests with serval state-of-the-art LMMs, including Claude-Opus-4.6 \cite{anthropic_claude_sonnet_46}, GPT-5.2 \cite{openai_gpt5mini} and Qwen3.5-VL \cite{bai2025qwen3vl}.
These tests assessed performance on schematic diagram understanding tasks, such as object detection, topological connectivity reasoning and spatial layout interpretation. 
As illustrated in Fig. \ref{fig:lmm_test}, each model can fail one of the schematic diagram understanding tasks. 
These failures indicate that current LMMs still struggle to jointly achieve precise visual grounding and topology-consistent reasoning, which constrains their effectiveness in schematic-to-netlist generation tasks requiring accurate textual grounding of schematic entities, faithful reconstruction of connectivity, and geometric reasoning over layout-dependent spatial relationships within schematic diagrams.
Recent benchmarks, such as Netlistfy \cite{huang2025netlistify}, MASALA-CHAI \cite{bhandari2024masala}, and PCB-Bench \cite{lipcb}, have begun to assess LMM performance on schematic diagram understanding; however, they concentrate on simplified schematics with few entities such as Simulation Program with Integrated Circuit Emphasis (SPICE) style circuits underscoring the need for benchmarks that enable more comprehensive evaluation of LMMs.

To bridge this gap, we introduce \textbf{\name}, a large-scale benchmark of real-world schematic diagrams that enables varied evaluation of LMMs across a diverse set of multimodal schematic understanding tasks.
As shown in Fig. \ref{fig:dataset_overview}, {\name} provides 1,854 high-quality real-world schematic designs with xx annotations of 109.9\textit{K} symbols (3.4\textit{K} unique types) and 245.4\textit{K} pins (with bounding boxes and semantic attributes), together with 423.4\textit{K} text instances and 219.8\textit{K} net graphs with  spatially weights and attributes.
We first propose structured evaluation protocols to assess netlist graph generated by LMMs. 
Our benchmark assesses four core abilities, including 
\textbf{(i) visual grounding} of schematic entities and texts,
\textbf{(ii) diagram-to-graph reasoning} for electrical connectivity extraction, and
\textbf{(iii) geometry-aware spatial reasoning} for constructing spatially weighted netlist graphs.
\textbf{(iv) tool-augmented agentic visual search} for query-specified circuit entities (components, connections, and graphs) within a schematic diagram.
In summary, the contributions of this work three-fold:

$\bullet$ \textbf{\name}: We present {\name}, a comprehensive benchmark for evaluating LMMs on schematic-to-graph reasoning, providing 1,854 real-world schematic designs with fine-grained annotations of symbols, pins, text instances, and spatially weighted net graphs, together with structured evaluation protocols.

$\bullet$ \textbf{Systematic Model Evaluation.}: We systematically evaluate state-of-the-art LMMs under varied settings, from single-shot prompting to tool-augmented agentic workflow, to quantify the impact of active visual search on schematic-to-netlist graph generation across diverse schematic understanding tasks.

$\bullet$ \textbf{Tool-augmented Agentic Framework}: We provide unified task formats, evaluation  protocols to support reproducible research and standardized comparison in this emerging field.

\input{comp_table2}

%% file: comp_table2.tex
\begin{table*}[t]
\centering
\caption{
Comparison of publicly available schematic-to-netlist benchmarks and datasets.
Abbreviations: 
Sch.—schematic diagram; 
Sp.Netlist—spatially weighted netlist.
}
\label{tab:pcb_benchmark}

\begingroup
\fontsize{5.5}{6.5}\selectfont
\setlength{\tabcolsep}{3pt}
\renewcommand{\arraystretch}{1.15}

\begin{tabular}{
>{\centering\arraybackslash}m{1.8cm}
>{\centering\arraybackslash}m{0.65cm}
>{\centering\arraybackslash}m{0.65cm}
>{\centering\arraybackslash}m{0.65cm}
>{\centering\arraybackslash}m{0.65cm}
>{\centering\arraybackslash}m{0.65cm}
>{\centering\arraybackslash}m{0.9cm}
>{\centering\arraybackslash}m{1.0cm}
>{\centering\arraybackslash}m{0.6cm}
>{\centering\arraybackslash}m{0.6cm}
>{\centering\arraybackslash}m{0.9cm}
>{\centering\arraybackslash}m{0.5cm}
}

\toprule

\multirow{2}{*}{Dataset} &
\multirow{2}{*}{Images} &
\multirow{2}{*}{\makecell{Avg. \\Sym.\\ Img}} &
\multirow{2}{*}{\makecell{Symbol\\Cat.}} &
\multirow{2}{*}{BBox} &
\multirow{2}{*}{\makecell{Spatial\\Netlist}} &
\multirow{2}{*}{\makecell{EDA\\Render\\Engine}} &
\multicolumn{4}{c}{Annotation} &
\multirow{2}{*}{Year} \\

\cmidrule(lr){8-11}

& & & & & & &
Symb. & Pin & Text & Net & \\

\midrule

CGHD~\cite{thoma2021public}
& 2,424 & 42 & 45 & \cmark & \xmark & \xmark
& \cmark & \xmark & \cmark & B-Net
& 2021 \\

AMSNet~\cite{tao2024amsnet}
& 894 & 10 & 12 & \xmark & \xmark & \xmark
& \cmark & \xmark & \xmark & B-Net
& 2024 \\

Masala-CHAI~\cite{bhandari2024masala}
& 7,500 & 12 & 12 & \xmark & \xmark & \xmark
& \cmark & \xmark & \xmark & B-Net
& 2024 \\

AMSNet 2.0~\cite{shi2025amsnet20largeams}
& 2,686 & 10 & 12 & \xmark & \xmark & \xmark
& \cmark & \xmark & \xmark & B-Net
& 2025 \\

Image2Net~\cite{xu2025image2net}
& 2,914 & 15 & 22 & \xmark & \xmark & \xmark
& \cmark & \xmark & \xmark & B-Net
& 2025 \\

MuaLLM~\cite{abbineni2025muallm}
& 2,914 & 29 & 22 & \xmark & \xmark & \xmark
& \cmark & \xmark & \xmark & B-Net
& 2025 \\

Netlistify~\cite{huang2025netlistify}
& 100,000 & 11 & 16 & \cmark & \xmark & \xmark
& \cmark & \xmark & \xmark & B-Net
& 2025 \\

SINA~\cite{aldowaish2026sina}
& 662 & -- & 10 & \cmark & \xmark & \xmark
& \cmark & \xmark & \xmark & SW-Net
& 2026 \\

\rowcolor{rowblue}
\textbf{{\name} (ours)}
& \textbf{1,854} & \textbf{59} & \textbf{3,480}
& \cmark & \cmark & \cmark
& \cmark & \cmark & \cmark & SW-Net
& 2026 \\

\bottomrule
\end{tabular}

\endgroup
\end{table*}

%% file: related_work.tex
\section{Related Work}

\subsection{Visual Grounding}

LMMs such as Kosmos-2, Ferret, GLMM and LLaVA-Grounding \cite{peng2023kosmos2,you2023ferret,ma2024groma,rasheed2024glamm,liu2023llava} and  have rapidly moved beyond global image–text alignment to produce verifiable region-level outputs with location-token or box-annotated generation, which enables phrase grounding and referring-expression comprehension within a single instruction. 
Current visual grounding methods \cite{peng2023kosmos2,you2023ferret,ma2024groma,rasheed2024glamm} typically follow either end-to-end architectures or referential grounding formulations.
Concurrently emerging benchmarks \cite{zeng2024compositional,ma2024gigagrounding,zhao2025rgbtground} increasingly emphasize challenging settings such as dense visual clutter and compositional reasoning, aiming to better expose the remaining limitations of current approaches.
More recently, agentic visual search has emerged as a complementary paradigm for visual grounding, framing the task as an iterative decision-making process, such as MM-ReAct, DeepEyes, and V* \cite{yang2023mmreact,zheng2025deepeyes,wu2024v}.
Despite strong progress in visual grounding, LMMs grounding remains unsolved problem in certain important areas such as fine-grained localization under crowding, structured diagrams and relational referring.

\subsection{Benchmarks for Schematic-to-Netlist}

In recent years, schematic-to-netlist conversion has advanced alongside several public datasets and evaluation benchmarks. 
AMSNet \cite{tao2024amsnet} provides 894 analog mixed-signal (AMS) schematic diagrams paired with SPICE netlists, and AMSnet 2.0 \cite{shi2025amsnet} further scales this direction to 2,686 circuits by adding Spectre-formatted netlists, OpenAccess digital schematics, and positional annotations. 
Netlistify \cite{huang2025netlistify} releases a modular deep-learning pipeline and a 100,000-image dataset with component annotations (e.g., bounding boxes and orientations) with predefined labels for net extraction, targeting HSPICE-compatible netlist reconstruction. 
Beyond printed analog schematics, CGHD \cite{thoma2021public} provides hand-drawn circuits with 2,424 images. 
More recently, PCB-Bench \cite{lipcb} broadens the scope to PCB placement and routing and design comprehension, including roughly 3,700 QA instances with 174 schematic designs.
Current datasets focusing on SPICE-style schematic diagrams, which typically contain a limited number of entity types and samples, compared with thousands of components types in practical schematic designs.
Similarly, current methods remain confined to recognizing a narrow predefined set of predefined symbols under fixed rules, lacking the multimodal reasoning required to interpret authentic schematic design artifacts.
Due to limited space, more detailed information is provided in Appendix A.

%% file: benchmark.tex
\section{Benchmark Construction}

In this section, we describe the dataset description, annotation curation, statistics, and evaluation criteria. Due to space limitations, more details can be found in the Appendix~B.

\subsection{Task Formulation}

\begin{figure}[t]
    \centering
    \includegraphics[width=\linewidth]{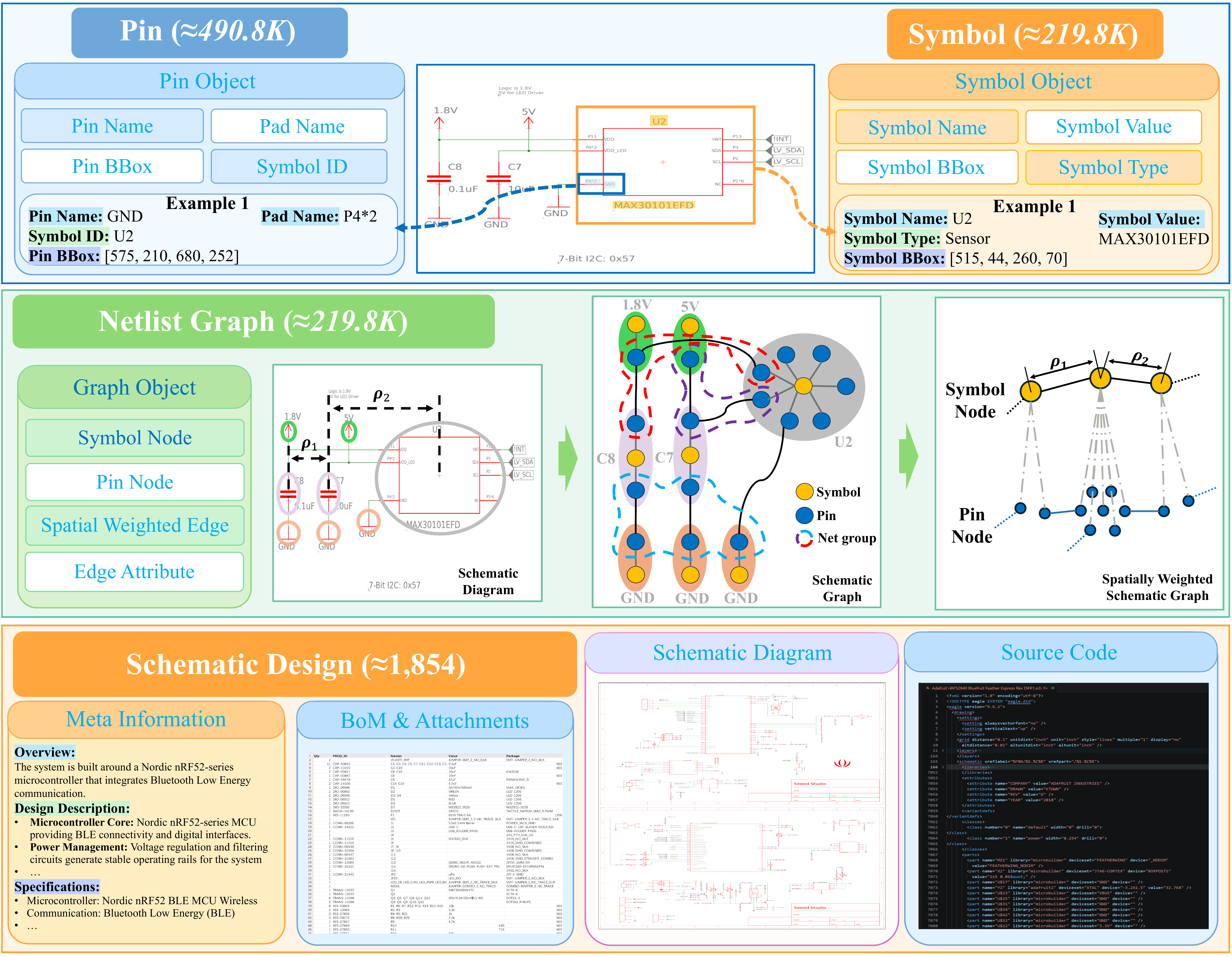}
    \caption{Overview of {\name} benchmark with representative cases.}
    \label{fig:dataset_overview}
\end{figure}

To provide a comprehensive evaluation framework for schematic-to-netlist generation tasks, we categories.
Detailed descriptions of these core capabilities are as follows.

$\bullet$ \textbf{Instance Detection.} Accurately localizing electronic components and pins is fundamental to schematic understanding. This capability is evaluated through symbol detection and pin detection tasks, where models must produce precise bounding boxes in dense, symbol-rich technical diagrams.

$\bullet$ \textbf{Symbol Classification.} Recognizing component types and their orientations is essential for establishing the functional identity of circuit elements. This capability is assessed through symbol type classification and orientation recognition tasks, requiring domain-specific visual discrimination under diverse stylistic variations.

$\bullet$ \textbf{Text Referring.} Schematics contain densely distributed text serving distinct semantic roles — symbol names, parametric values, net labels, and pin annotations. The ability to ground each text instance to its associated symbol or pin is evaluated through text-to-symbol linking and semantic role disambiguation tasks.

$\bullet$ \textbf{Topological Relation Reasoning.} Understanding electrical connectivity from visual layout demands multi-hop structural reasoning beyond local perception. This capability is assessed through pin–symbol membership prediction and pin–pin connectivity prediction tasks, where models must infer how components are electrically connected from spatial context.

$\bullet$ \textbf{Heterogeneous Graph Construction.} Beyond pairwise connectivity understanding, converting an entire schematic into a unified structured representation requires holistic diagram parsing. This capability is evaluated by requiring models to generate a complete spatially weighted heterogeneous graph — comprising symbol nodes, pin nodes, membership edges, and net-induced connectivity edges — that faithfully encodes both the hierarchical composition and global topology of the circuit in a single pass.

\begin{figure}[t]
    \centering
    \includegraphics[width=\linewidth]{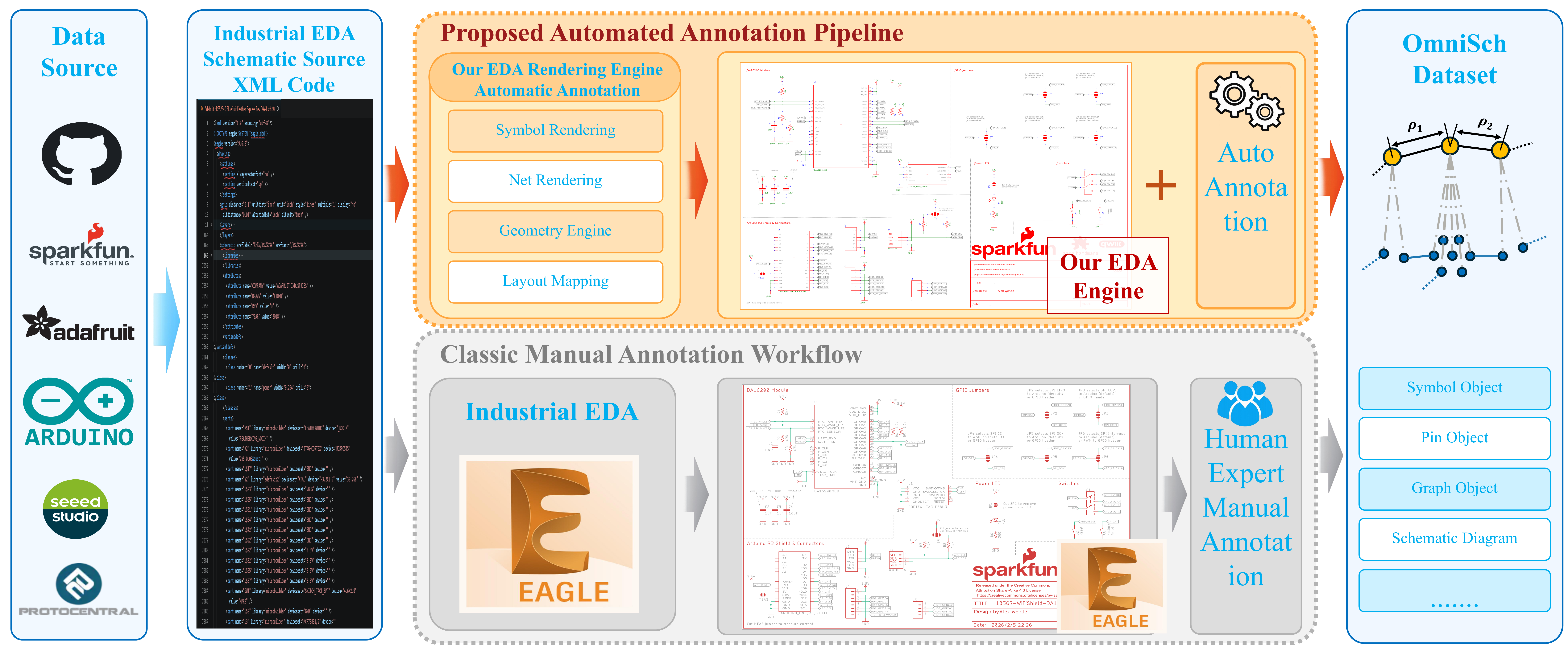}
    \caption{\textbf{Comparison between different data annotation paradigms. (a) Manual annotation:}. Relying on human expert on labeling . \textbf{(b) Auto annotation:} Our {\name} directly directly render schematic source code and automatically label data via our EDA rendering engine. }
    \label{fig:dataset_pipeline}
\end{figure}

\subsection{Annotation Curation}

The {\name} benchmark is constructed through a carefully designed pipeline for data collection and processing, as illustrated in Figure~\ref{fig:dataset_pipeline}.
We began by collecting real-world schematics from multiple open-source communities, such as SparkFun, Arduino, Adafruit, and GitHub \cite{sparkfun, github, arduino, adafruit, seeedstudio, protocentral}.
Based on these sources, we curate schematic design files created with the industrial EDA tool Autodesk EAGLE \cite{autodesk_eagle}, because its XML-based schematic format, enabling scalable extraction of diagram instances and connectivity for automated annotation.

Unlike traditional workflows that rely on time-consuming, costly, and labor-intensive manual annotation (Figure ~\ref{fig:dataset_pipeline}), we propose an automated pipeline via our custom EDA generative rendering engine.
Our EDA generative rendering engine parses EAGLE XML source files and re-simulates the native schematic rendering process to produce high-fidelity schematic images, while simultaneously exporting pixel-aligned annotations at multiple granularities—from fine-grained entity localization and semantic labeling (e.g., symbol/pin bounding boxes and attributes) to net-level connectivity represented as spatially weighted netlist graphs.
Due to space limitation, the implementation details can be found in Appendix~B.

\begin{figure*}[h]
    \centering
    \def\figHeight{2cm} 

    \subcaptionbox{20 Topic distribution.\label{fig:stats_topics}}%
    {\includegraphics[height=\figHeight]{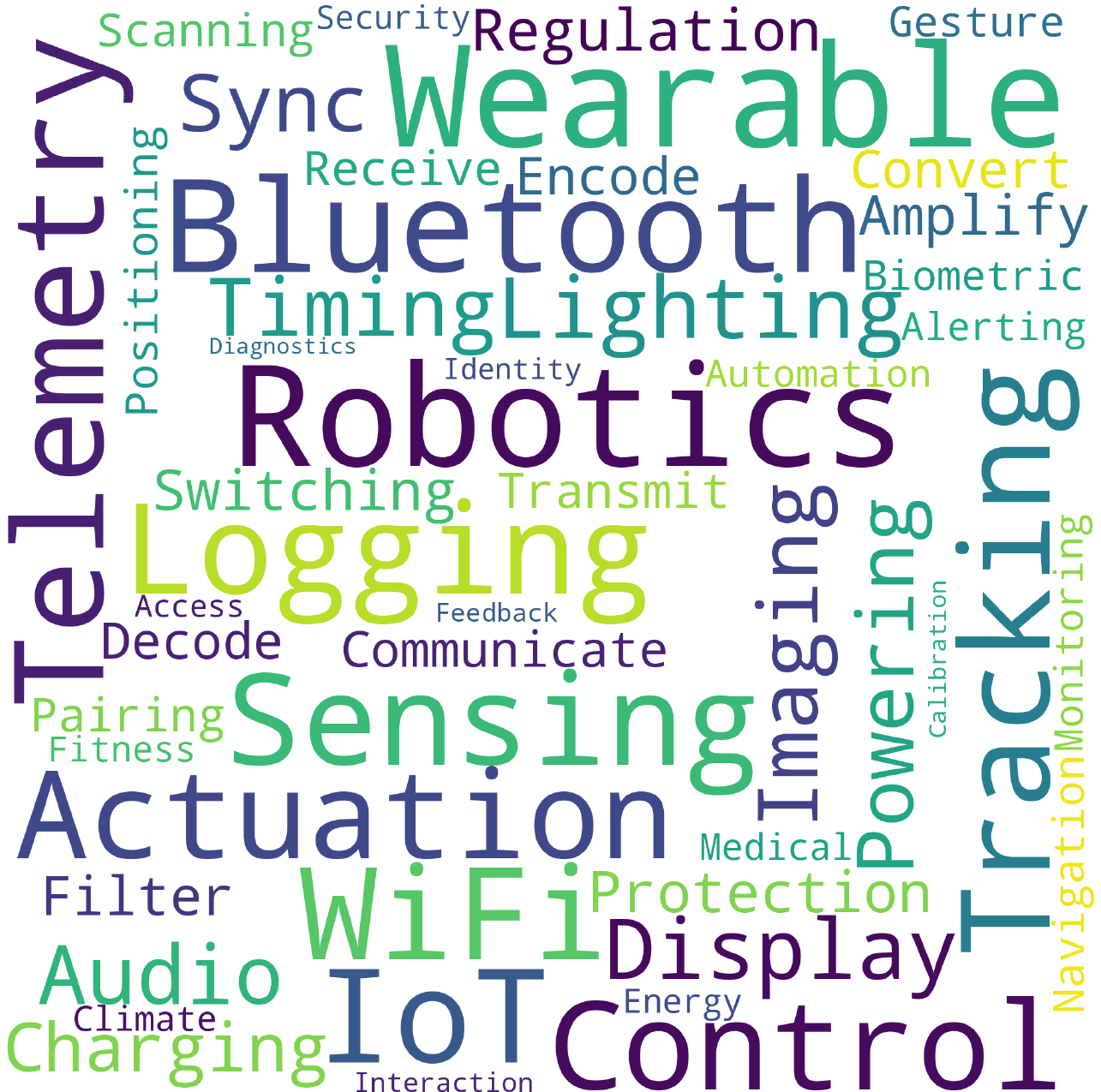}}%
    \hfill 
    \subcaptionbox{Symbols per image.\label{fig:stats_symbols}}%
    {\includegraphics[height=\figHeight]{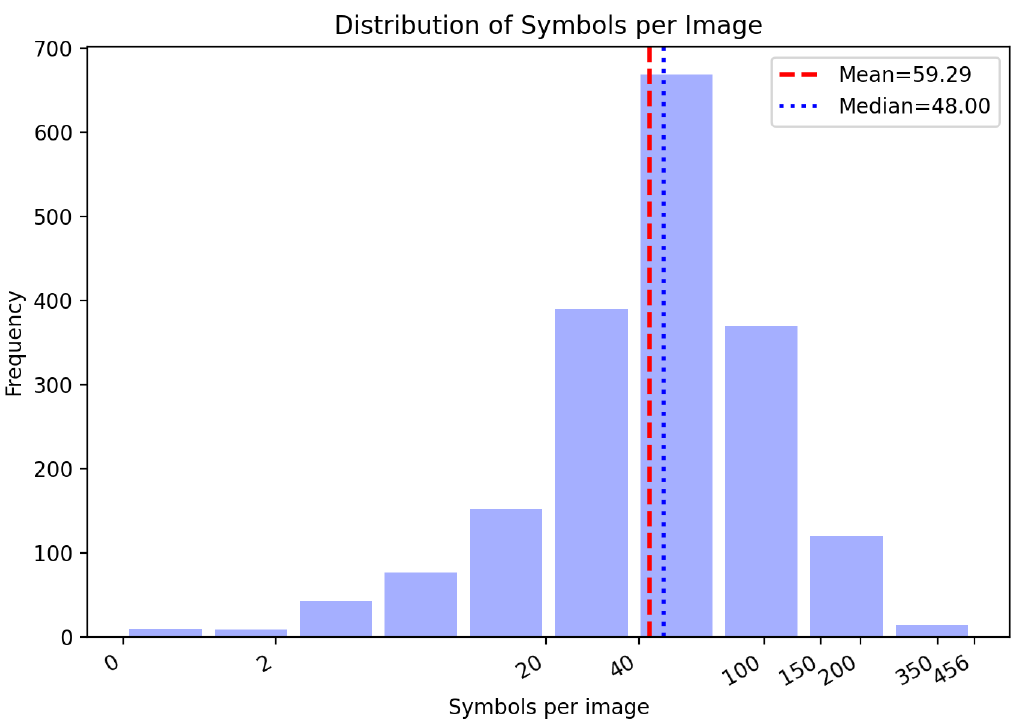}}%
    \hfill 
    \subcaptionbox{Pins per image.\label{fig:stats_pins}}%
    {\includegraphics[height=\figHeight]{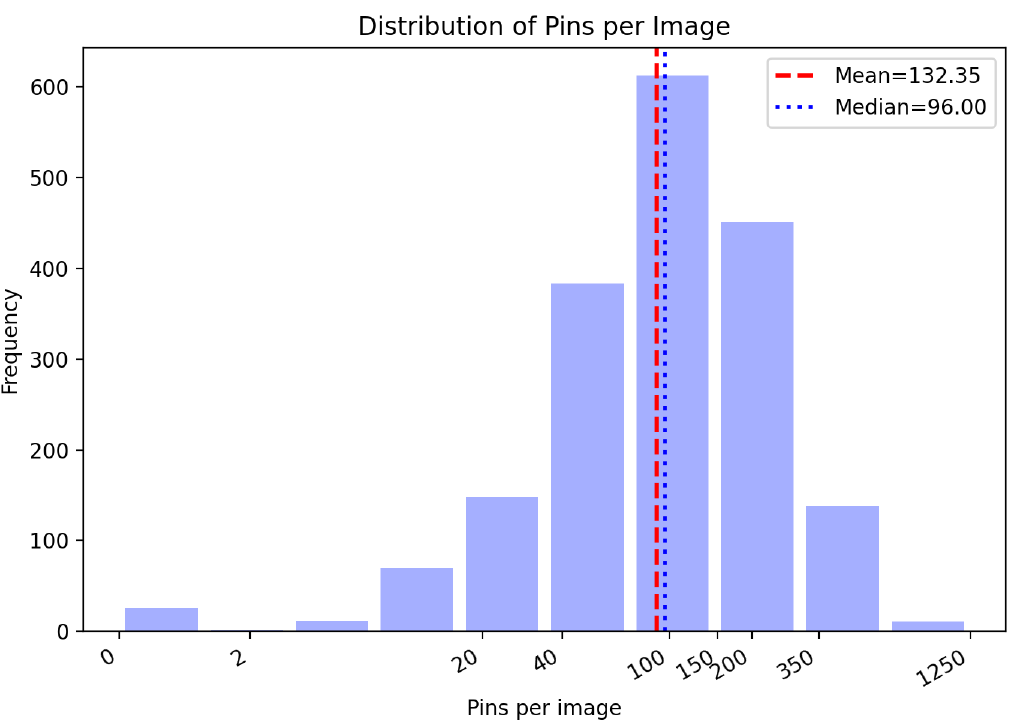}}%
    \hfill 
    \subcaptionbox{Nets per image.\label{fig:stats_nets}}%
    {\includegraphics[height=\figHeight]{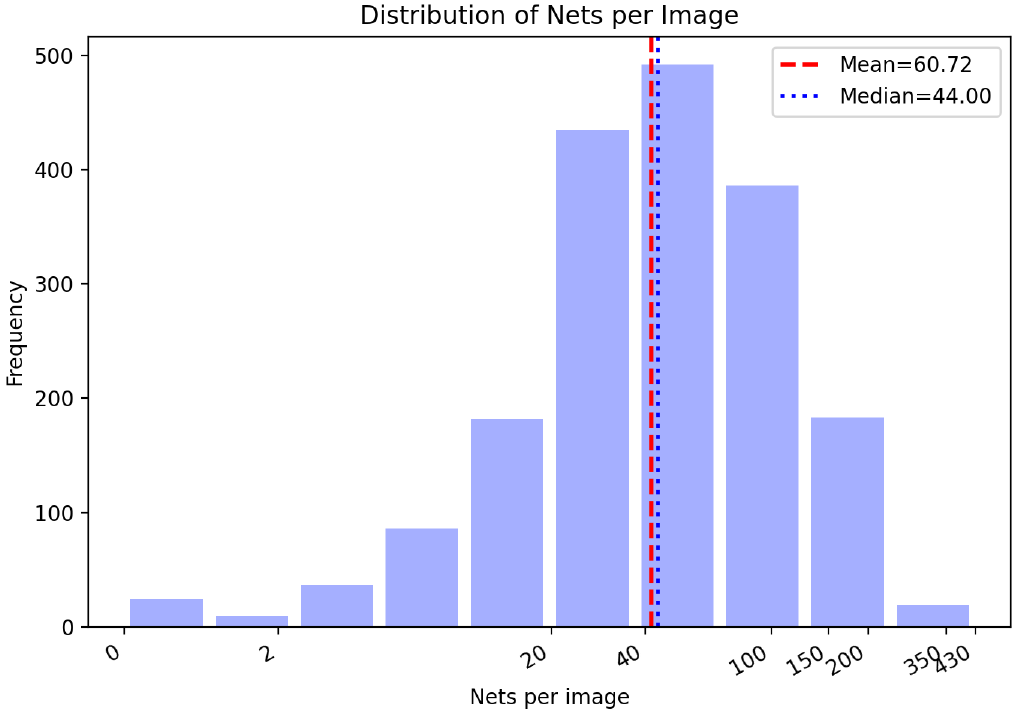}}%


    \caption{Statistical overview of the {\name} benchmark. The dataset encompasses a diverse range of electronic domains, comprising 1-440 symbols, 1-1200 pins, 1-400 nets, and 1-1600 text instances. This large-scale diversity provides a comprehensive benchmark for the automatic generation and evaluation of schematic netlists.}
    \label{fig:img2sch_statistic}
    
\end{figure*}

\subsection{Statistics of OmniSch Benchmark} 
As illustrated in Figure~\ref{fig:img2sch_statistic} and Figure~\ref{fig:dataset_overview}, {\name} comprises 1,854 real-world schematic diagrams spanning diverse application domains (e.g., robotics, wireless, and sensing), providing a rich resource for supervised training and fine-tuning on practical schematic understanding and EDA tasks.

$\bullet$ \textbf{Schematic Diagram Instance.} 
We construct a fine-grained instance-level annotation set that localizes and semantically describes the visual elements in a schematic, including symbols, pins, and texts, as illustrated in Figure \ref{fig:dataset_overview}. For each symbol instance, we annotate its bounding box, symbol type, orientation, and semantic attributes (symbol name and symbol value). For each pin instance, we annotate its bounding box, its parent symbol ID, and pin-level attributes (pin name and pad name). For each text instance, we annotate its location, its content, and its reference target.

$\bullet$ \textbf{Connectivity Annotations.} 
We construct topological net-level connectivity annotations by explicitly annotating pin–pin edges that indicate which pins are electrically connected in the schematic. Each edge is augmented with a spatial weight computed as the diagonal distance between the two parent symbols that the connected pins belong to, normalized by the schematic image size. When a connection is labeled in the drawing, we additionally annotate the corresponding net name as an attribute of the same edge.

$\bullet$ \textbf{Schematic graphs.}
We construct a heterogeneous, spatially weighted schematic graph that unifies instances and connectivity into a single representation. The graph contains symbol nodes and pin nodes, includes membership edges linking each pin to its parent symbol, and includes net-induced connectivity edges between pins (and optionally derived symbol links), each associated with net attributes and spatial weights. 
This representation preserves the global topology while explicitly retaining local clustering structure that reflects functional blocks as represented in the schematic drawing.

\subsection{Evaluation Criteria}

We adapt seven types of evaluation metrics tailored to specific task categories. In the following, we present an overview of the evaluation metrics and their applicability to specific tasks.

$\bullet$ \textbf{Instance Detection Type.} To evaluate the model's ability to localize symbols and pins, we employ the F1 score to assess whether each ground-truth instance is successfully detected, without requiring precise spatial overlap at this stage.

$\bullet$ \textbf{Semantic Attribute Type.} For text referring and semantic attribute extraction, we adopt the F1 score with exact string matching to determine whether the grounded text precisely matches the ground-truth annotation. Approximate or similarity-based matching is deliberately excluded, as real-world schematic-to-netlist conversion demands rigorous character-level correctness in component designators and parametric values.

$\bullet$ \textbf{Localization Type.} For symbol and pin bounding box evaluation, the IoU score is applied to quantify the spatial overlap between predicted regions and ground-truth annotations, measuring the model's fine-grained localization precision.

$\bullet$ \textbf{Connectivity Type.} To evaluate topological relation reasoning, we employ the F1 score to assess the matching accuracy of predicted pin–pin connections against ground-truth nets, and additionally report accuracy to measure the recall coverage of detected nets across the full circuit.

$\bullet$ \textbf{Graph Structure Type.} For heterogeneous graph construction, we adopt the IoU score over edge sets to evaluate the structural overlap between the predicted graph and the ground-truth graph, capturing how faithfully the generated representation preserves both membership and connectivity edges.

$\bullet$ \textbf{Spatial Layout Type.} To assess whether the predicted graph preserves the relative spatial arrangement of the original schematic, we employ Kendall's Tau \cite{kendall1938new} rank correlation coefficient, measuring the consistency of pairwise spatial orderings between predicted and ground-truth node positions.

\subsection{Netlist Graph Matching Protocol}

To the best of our knowledge, there is no standardized protocol for matching netlist graphs predicted by LMMs with ground-truth netlist graphs.
To evaluate a predicted netlist graph $G_p=(V_p,E_p)$ against the ground-truth graph $G_g=(V_g,E_g)$, we consider node sets composed of symbol nodes and pin nodes, i.e., $V_g = V_g^{\mathrm{sym}} \cup V_g^{\mathrm{pin}}$. The graph has a hierarchical structure in which symbol nodes represent higher-level components, while pin nodes are dependent entities associated with symbols, and connectivity is defined through pin participation rather than direct symbol–symbol edges. The edge set contains only symbol--pin edges and pin--pin edges, i.e., $E_g = E_g^{\mathrm{sym\mbox{-}pin}} \cup E_g^{\mathrm{pin\mbox{-}pin}}.$
Under this formulation, direct graph comparison is challenging because predicted and ground-truth graphs lack shared node identities, and IoU at the geometric level is unreliable due to imprecise bounding box predictions from LMMs. 
In order to enable meaningful evaluation of edges, attributes, and overall topology, we first establish node correspondence via bipartite matching and then perform all subsequent comparisons under the induced mapping.
The matching equations are following:


\noindent\textbf{Symbol matching equation}
For symbol nodes, the matching score is defined as
\begin{equation}
\begin{aligned}
S_{\mathrm{sym}}(\hat{u},u)
&=
w_t\,\mathbf{1}[\hat{T}_{\mathrm{sym}} = T_{\mathrm{sym}}]
+ w_n\,S_{\mathrm{txt}}(\hat{n}, n)
+ w_v\,S_{\mathrm{txt}}(\hat{v}, v)
+ w_r\,S_{\mathrm{pos}}(\hat{r}, r),
\end{aligned}
\label{eq:sym_unified}
\end{equation}
where $\hat{u}$ and $u$ denote a predicted symbol node and a ground-truth
symbol node, respectively; $\hat{T}_{\mathrm{sym}}$ and $T_{\mathrm{sym}}$
denote their symbol class indices (defined in the Appendix); $\hat{n}$ and
$n$ denote the symbol names; $\hat{v}$ and $v$ denote the symbol values;
$\hat{s}$ and $s$ denote degree-based structural signatures; and $\hat{r}$
and $r$ denote relative-position signatures. Here,
$S_{\mathrm{txt}}(\cdot,\cdot)$ measures text similarity, and
$S_{\mathrm{pos}}(\cdot,\cdot)$ measures agreement in coarse relative position.

\noindent\textbf{Pin matching equation}
For pin nodes, the matching score is defined as
\begin{equation}
\begin{aligned}
S_{\mathrm{pin}}(\hat{q},q)
&=
 w_n\,S_{\mathrm{txt}}(\hat{n}, n)
+ w_p\,S_{\mathrm{txt}}(\hat{p}, p)
+ w_a\,S_{\mathrm{txt}}(\hat{a}, a)
+ w_c\,S_{\mathrm{comp}}(\hat{c}, c),
\end{aligned}
\label{eq:pin_unified}
\end{equation}
where $\hat{q}$ and $q$ denote a predicted pin node and a ground-truth pin
node, respectively; $\hat{g}$ and $g$ denote their optional segment
identifiers; $\hat{n}$ and $n$ denote the net names; $\hat{p}$ and $p$
denote the pin names; $\hat{a}$ and $a$ denote the pad names; and $\hat{c}$
and $c$ denote coarse structural signatures, such as the sizes of the
corresponding connected components in the pin--pin graph. Here,
$S_{\mathrm{txt}}(\cdot,\cdot)$ measures text similarity, and
$S_{\mathrm{comp}}(\cdot,\cdot)$ measures coarse structural compatibility.

Given the pairwise similarity matrix constructed from Eqs.~\eqref{eq:sym_unified} and Eqs.~\eqref{eq:pin_unified}, correspondence is established in a hierarchical manner. We first compute one-to-one matching between $V_p^{\mathrm{sym}}$ and $V_g^{\mathrm{sym}}$ via maximum-weight bipartite matching. Subsequently, for each aligned symbol pair, we match the associated $(V_p^{\mathrm{pin}},V_g^{\mathrm{pin}})$ under the dependent-symbol constraint. Based on the resulting node correspondence, we then evaluate net-level attributes, edge connectivity, and overall graph consistency. Due to space limitations, further implementation details are provided in the \tai{Appendix D}.

\subsection{Evaluation Criteria}

We define seven types of evaluation metrics tailored to different task categories, each capturing a specific aspect of schematic understanding. Our evaluation metrics include: \textbf{(i)} \textbf{Instance Detection}, which uses F1 score to assess whether ground-truth symbol and pin instances are successfully detected, without requiring precise spatial overlap; 
\textbf{(ii)} \textbf{Semantic Attributes}, which evaluates text referring and attribute extraction using F1 score with exact string matching, enforcing strict correctness for symbol names, values, and pin/pad labels; 
\textbf{(iii)} \textbf{Localization}, which applies IoU to measure the spatial accuracy of predicted symbol and pin bounding boxes; 
\textbf{(iv)} \textbf{Connectivity}, which uses F1 score to evaluate predicted pin–pin connections against ground-truth nets, along with accuracy to measure coverage of circuit connectivity; 
\textbf{(v)} \textbf{Graph Structure}, which adopts IoU over edge sets to quantify structural overlap between predicted and ground-truth graphs; and 
\textbf{(vi)} \textbf{Spatial Layout}, which employs Kendall’s Tau~\cite{kendall1938new} to measure consistency in relative spatial ordering between predicted and ground-truth node positions.
\tai{\textbf{(vii)} \textbf{Agentic Tool Use}, which measures using (a) \textit{Target Coverage}, defined as the maximum IoU between queried regions and the ground-truth target to assess localization accuracy; (b) \textit{Step Efficiency}, defined as the average number of steps required to reach a correct answer; and (c) \textit{Trace-following Quality}, which evaluates whether the sequence of queries correctly follows circuit connectivity to infer topology-dependent information such as net names.
}

%% file: experiment.tex
\section{Experiment and Findings}

\begin{figure}[h]
    \centering
    \includegraphics[width=\linewidth]{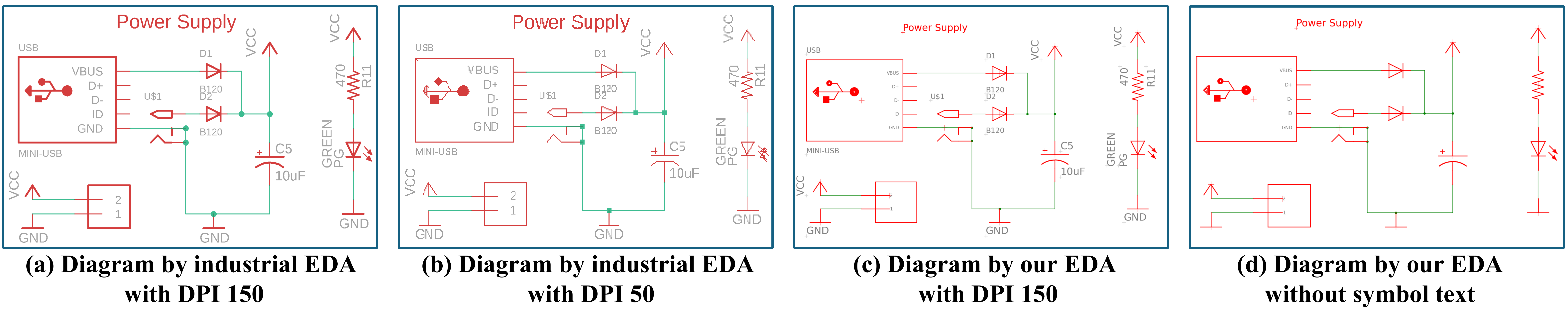}
    \caption{Synthesized schematic variations generated by our custom EDA generative rendering engine. The first diagram shows the original export from the industrial EDA tool; all remaining diagrams are rendered by our engine under controlled variations. (a) original EDA export; (b) full text; (c) without all text; (d) without symbol names and values.}
    \label{fig:synth_data}
\end{figure}

\subsection{Experimental Setups}

\textbf{Study Setup.} The tested LMMs in this section include GPT5.2~\cite{openai_gpt5}, GPT-5-mini~\cite{openai_gpt5}, Claude-Sonnet-4.6~\cite{anthropic_claude_sonnet_46}, Gemini-2.5-Flash-Lite~\cite{google_gemini_25_flash}, Gemini-3.1-Pro-Preview~\cite{google_gemini_31_pro}, Qwen3-8B-Instruct~\cite{bai2025qwen3vl}, Qwen3-VL-235B-A22B~\cite{bai2025qwen3vl}, Ministral-14B~\cite{ministral}, and LLaMA-4-Maverick-400B~\cite{meta_llama4}.
We evaluate these models alongside a classical vision pipeline baseline and our proposed agentic framework, as described below.
For training the classical pipeline components (e.g., object detectors), we randomly split {\name} into 80\% training and 10\% validation sets, with the remaining 10\% reserved as a held-out test set for all models.
We conduct three studies: \textbf{(i)} \textbf{Zero-shot Evaluation}, where all LMMs are tested using a single prompt without intermediate interaction to assess their end-to-end ability to convert schematic images into netlist graphs. We also include a domain-specific SFT-trained model based on Qwen3-8B-Instruct \cite{bai2023qwen}, along with a classical vision pipeline (discussed below), for comparison; \textbf{(ii)} \textbf{Ablation on Synthesized Schematics}, where we evaluate model robustness on schematics generated by our EDA rendering engine under controlled variations (text configurations, resolution, and color), as shown in Figure~\ref{fig:synth_data}. We also conduct few-shot experiments to assess the impact of example schematics on performance; and \tai{\textbf{(iii)} \textbf{Agentic Evaluation}}, where we evaluate the impact of active visual search on schematic-to-netlist graph generation using a unified tool-augmented framework (described below) under a zero-shot setting.
Due to space limitation, full implementation details are provided in Appendix D.

\textbf{Implementation of Classical Vision Baseline.} We construct a classical, modular vision pipeline as a principled baseline for evaluating LMMs on schematic understanding, providing a transparent reference to diagnose where end-to-end models succeed or fail. The pipeline relies on YOLO11~\cite{yolo11} for visual perception and PaddleOCR~\cite{paddleocr3_2024} for text extraction, followed by a series of rule-based and algorithmic modules to construct symbol, pin, and net instances and recover circuit connectivity. 
For text referring, we develop a multimodal text grounding model that associates OCR tokens with schematic instances to extract symbol names, values, and pin/pad labels.
These structured outputs are then reconciled into a global netlist.

\begin{figure}[H]
    \centering
    \includegraphics[width=\linewidth]{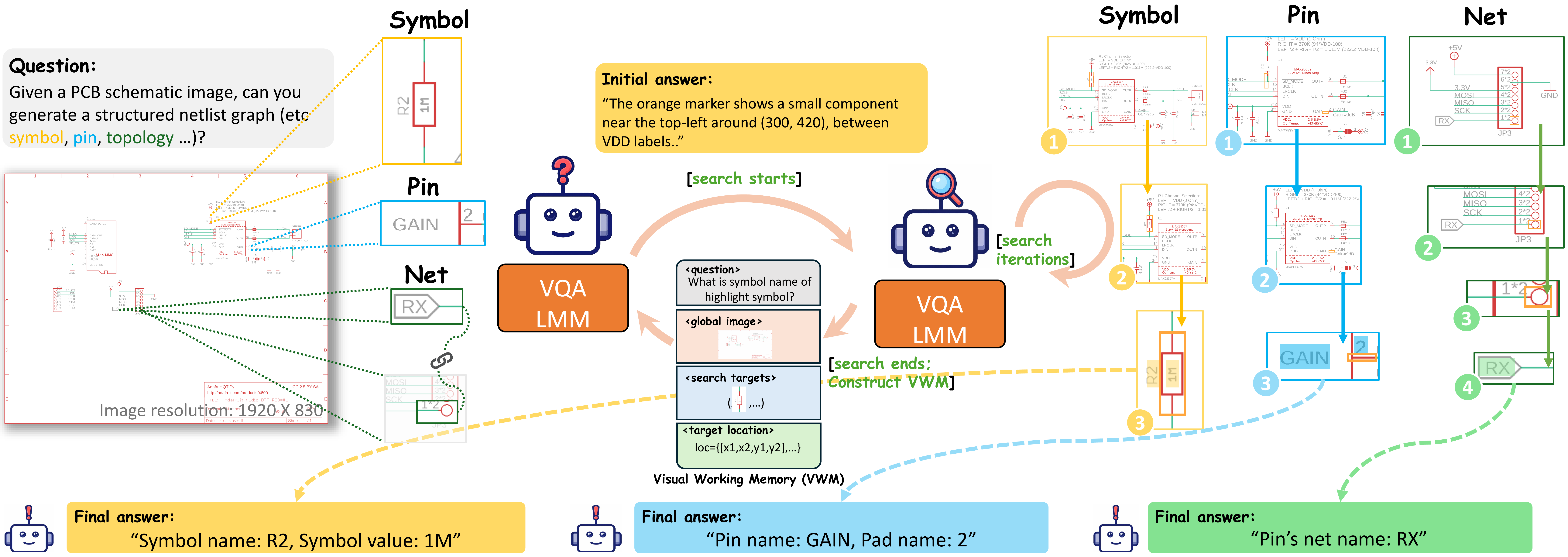}
    \caption{Overview of the ReAct-based agentic framework for LMMs to evaluate usage and performance of LMMs using tools in schematic-to-netlist 
conversion.}
    \label{fig:react_framework}
\end{figure}


\textbf{Implementation of LMMs Agentic Framework.} We design a React-based evaluation framework to benchmark how LMMs perform VQA-style reasoning to parse PCB schematic images into structured netlists, with a focus on text referring tasks such as symbol names, component values, pin names, pad names, and net names. The framework presents a high-resolution schematic and a structured query, and asks the LMMs to iteratively issue crop operations to localize target regions and refine its predictions through a search-and-zoom process.

\subsection{Main Results and Findings}

\textbf{Evaluation results of zero-shot study} are shown in Table~\ref{tab:task1_new}. 
Overall, the graph construction performance of LMMs is significantly lower than that of human engineers (last row), indicating their limited capability in comprehensive schematic understanding. Evaluation results on schematic understanding tasks reveal consistent limitations of LMMs across multiple capabilities.

\textbf{$\bullet$ Instance detection.} LLMs struggle in dense diagrams, particularly for fine-grained elements such as pins, and rely heavily on textual cues rather than true visual grounding, with accuracy dropping noticeably when textual information is removed.
Furthermore, LMMs exhibit limited capability in precise visual grounding. For example, although Gemini 2.5 Flash achieves \tai{66.36\%} accuracy in symbol detection under match by Eq.~(\ref{eq:sym_unified}), its IoU-based matching score is only \tai{0.00\%}. This gap suggests that while LMMs can roughly identify instances—often leveraging textual information—they fail to accurately localize their precise spatial boundaries.

\textbf{$\bullet$ Symbol Classification.} LMMs can identify basic component types (e.g., resistors and capacitors) but fail to generalize to more complex or functional symbols. 

\textbf{$\bullet$ Text Referring.} LMMs exhibit substantial weaknesses, often failing to accurately associate textual information (e.g., net names, symbol names, and values) with corresponding visual regions, highlighting limitations in precise grounding. 

\textbf{$\bullet$ Topological Relation Reasoning.} LMMs still struggle to understand connectivity among schematic instances. From the net perspective, performance remains low across models: on real schematics, even the best LMM reaches only $27.44\%$ net detection F1 and $25.07\%$ net-name accuracy. This suggests that current systems remain weak at recovering long-range circuit connectivity and assigning semantic identities to nets. Moreover, performance further degrades as the number of nets increases, highlighting the difficulty of reasoning over more complex connectivity patterns; we discuss this trend further in the appendix.

\textbf{$\bullet$ Geometric Relation Understanding.} Table~\ref{tab:task1_new} suggests that current LMMs still exhibit limited performance in spatially grounded visual understanding. On real schematics, IoU-based symbol detection for all LMMs remains below $0.3\%$, and Kendall’s $\tau$ stays below 0.05, indicating that even when the models can recognize the correct symbols semantically, they still struggle to ground them accurately in image space and to preserve relative geometric relationships

\textbf{$\bullet$ Heterogeneous Graph Construction.} These deficiencies propagate to overall graph generation, where overall graph quality remains low, as reflected by poor IoU and high GED scores. Together, these results demonstrate that current LMMs lack robust capabilities in fine-grained perception, and structural reasoning required for accurate schematic understanding.

\input{eval_pipeline1}

\textbf{Evaluation result on ablation study} are shown in 
Table \ref{tab:task1_new}.
The results of this ablation study reveal how LMM performance varies under controlled visual conditions and demonstrate the extent to which factors such as resolution, text availability, and few-shot examples influence robustness and overall understanding.

\textbf{$\bullet$ Impact of Text Configurations.} The effect of text removal is not uniform across models. For some LMMs, the \textit{w/o Txt.} setting even yields higher detection F1 than \textit{Synth.}, suggesting that text can sometimes interfere with recognition. For example, symbol detection F1 increases from $62.47\%$ to $72.35\%$ for Claude Sonnet 4.6 and from $56.22\%$ to $72.41\%$ for Gemini 2.5 Flash. In these cases, precision drops slightly while recall increases more substantially because the models make more predictions under \textit{w/o Txt.}, recovering more true instances but also introducing more false positives.

\textbf{$\bullet$ Impact of Diagram Resolution and FewShot.}
\tai{LMMs exhibit further degradation when processing low-resolution images, as the loss of visual detail impairs structural understanding, leading to lower Graph IoU and higher Graph GED, and ultimately resulting in a significant decline in overall performance.}

\input{eval_task3_3}

\textbf{Evaluation results on agentic evaluation} are shown in Table \ref{tab:combined_results}. LMMs show limited capability in text-referred visual search, especially in the full-image setting where failures in target localization significantly hinder performance. When the target region is already known and does not need to be localized, performance improves consistently across all metrics.



\textbf{$\bullet$ Topology Trace Understanding.} 
\tai{LMMs remain unable to reliably perform trace-based visual search for circuit connectivity. Notably, Gemini 3.1 Pro achieves the strongest performance in trace following, demonstrating precise step-by-step tracking over xx steps and successfully covering the target connectivity.}

%% file: eval_pipeline1.tex
\begin{table*}[t]
\centering
\caption{\textbf{Evaluation reults of LMMs on single prompt zero-shot and Ablation Study schematic-to-netlist generation.} \textit{Real} denotes the schematic exported by industrial EDA; \textit{Synth.} denotes the synthesis schematic by our custom EDA; \textit{GED} denotes as graph edit distance. Symbol attributes (Name, Value, Type) are evaluated on matched symbol detections only, and pin attributes (PinName, PadName) are evaluated on matched pin detections only.}
\label{tab:task1_new}

\begingroup
\scriptsize
\setlength{\tabcolsep}{3pt}
\renewcommand{\arraystretch}{1.1}

\resizebox{\textwidth}{!}{%
\begin{tabular}{l c c ccccc cccc c c cc c c}
\toprule

\multirow{3}{*}{Model}
& \multirow{3}{*}{Size}
& \multirow{3}{*}{\shortstack{Image\\Input}}
& \multicolumn{5}{c}{Symbol}
& \multicolumn{4}{c}{Pin}
& \multicolumn{2}{c}{Net}
& \multicolumn{2}{c}{Graph}
& Geometric
& Pass Rate \\

\cmidrule(lr){4-8}
\cmidrule(lr){9-12}
\cmidrule(lr){13-14}
\cmidrule(lr){15-16}
\cmidrule(lr){17-17}
\cmidrule(lr){18-18}

& &
& \multicolumn{2}{c}{Detection (F1)}
& \multicolumn{3}{c}{Attribute}
& \multicolumn{2}{c}{Detection (F1)}
& \multicolumn{2}{c}{Attribute}
& \multicolumn{1}{c}{Detection}
& \multicolumn{1}{c}{Attribute}
& \multicolumn{2}{c}{Structure}
& \multicolumn{1}{c}{Structure}
& \\

\cmidrule(lr){4-5}
\cmidrule(lr){6-8}
\cmidrule(lr){9-10}
\cmidrule(lr){11-12}
\cmidrule(lr){13-13}
\cmidrule(lr){14-14}
\cmidrule(lr){15-16}
\cmidrule(lr){17-17}
\cmidrule(lr){18-18}
& &
& \makecell[c]{Match by\\Eq. (\ref{eq:sym_unified})} 
& \makecell[c]{Match by\\IoU}
& \makecell[c]{Name} 
& \makecell[c]{Value} 
& \makecell[c]{Type}
& \makecell[c]{Match by\\Eq. (\ref{eq:pin_unified})} 
& \makecell[c]{Match by\\IoU}
& \makecell[c]{PinName} 
& \makecell[c]{PadName}
& \makecell[c]{F1}
& \makecell[c]{NetName}
& \makecell[c]{IoU} 
& \makecell[c]{1-GED}
& \makecell[c]{Kendall's $\tau$}
& \makecell[c]{Pass} \\
\midrule

\rowcolor{gray!20}
\multicolumn{18}{c}{Commercial Chatbot Systems} \\
\midrule

\multirow{3}{*}{GPT-5.2} & - & Real
& 0.5276 & 0.0009 & 0.1761 & 0.2434 & 0.2948
& 0.5155 & 0.0003 & 0.1607 & 0.1780
& 0.1987 & 0.1822
& 0.1546 & 0.0945
& 0.0337 & 0.7097 \\
& & Real Low R.
& \graycell 0.3184 & \graycell 0.0032
& \graycell 0.0363 & \graycell 0.0566 & \graycell 0.0451
& \graycell 0.1756 & \graycell 0.0000
& \graycell 0.0641 & \graycell 0.0547
& \graycell 0.0612 & \graycell 0.0504
& \graycell 0.0480 & \graycell 0.0590
& \graycell 0.0170 & \graycell 0.9946 \\
& & Synth.
& 0.4732 & 0.0007
& 0.1460 & 0.2196 & 0.2600
& 0.5217 & 0.0003
& 0.2160 & 0.1738
& 0.2588 & 0.2445
& 0.1722 & 0.0944
& 0.0501 & 1.0000 \\
& & w/o Txt.
& \graycell 0.4508 & \graycell 0.0011
& \graycell --- & \graycell --- & \graycell 0.1229
& \graycell 0.5369 & \graycell 0.0005
& \graycell 0.1802 & \graycell 0.3896
& \graycell 0.3924 & \graycell 0.0538
& \graycell 0.0116 & \graycell 0.1413
& \graycell 0.0097 & \graycell 1.0000 \\
& & FewShot.
& 0.6196 &  0.0000
& 0.1890 & 0.2999 & 0.4185
& 0.5469 &  0.000
& 0.1058 & 0.1724
& 0.4199 & 0.1716
& 0.1450 & 0.1806
& 0.0019 & 1.0000 \\
\midrule

\multirow{3}{*}{GPT-5-mini}  & - & Real
& 0.2997 & 0.0010 & 0.1026 & 0.1111 & 0.1320
& 0.3658 & 0.0000 & 0.1233 & 0.0892
& 0.1016 & 0.0921
& 0.0704 & 0.0730
& 0.0192 & 0.9032 \\
& & Real Low R.
& \graycell 0.1926 & \graycell 0.0033
& \graycell 0.0292 & \graycell 0.0476 & \graycell 0.0306
& \graycell 0.1391 & \graycell 0.0000
& \graycell 0.0362 & \graycell 0.0298
& \graycell 0.0315 & \graycell 0.0280
& \graycell 0.0210 & \graycell 0.0431
& \graycell 0.0057 & \graycell 0.9946 \\
& & Synth.
& 0.2918 & 0.0010
& 0.0945 & 0.1086 & 0.1287
& 0.3346 & 0.0000
& 0.1125 & 0.0881
& 0.0994 & 0.0905
& 0.0651 & 0.0659
& 0.0180 & 1.0000 \\
& & w/o Txt.
& \graycell 0.2312 & \graycell 0.0009
& \graycell --- & \graycell --- & \graycell 0.0474
& \graycell 0.3131 & \graycell 0.0000
& \graycell 0.0565 & \graycell 0.1924
& \graycell 0.1926 & \graycell 0.0297
& \graycell 0.0045 & \graycell 0.0732
& \graycell 0.0207 & \graycell 1.0000 \\
& & FewShot.
& 0.5310 &  0.0007
& 0.1713 & 0.1212 & 0.2284
& 0.3812 &  0.0001
& 0.0817 & 0.0000
& 0.2428 & 0.0802
& 0.1062 & 0.1199
& 0.0162 & 1.0000 \\

\midrule

\multirow{3}{*}{Claude Sonnet 4.6}  & - & Real
& 0.5862 & 0.0009 & 0.1487 & 0.2205 & 0.3637
& 0.5513 & 0.0003 & 0.1073 & 0.2113
& 0.1748 & 0.1594
& 0.1241 & 0.1590
& 0.0301 & 0.9839 \\
& & Real Low R.
& \graycell 0.5130 & \graycell 0.0012
& \graycell 0.0475 & \graycell 0.1205 & \graycell 0.2735
& \graycell 0.4017 & \graycell 0.0002
& \graycell 0.0602 & \graycell 0.1356
& \graycell 0.1158 & \graycell 0.1048
& \graycell 0.0837 & \graycell 0.1360
& \graycell 0.0221 & \graycell 0.9946 \\
& & Synth.
& 0.6247 & 0.0024
& 0.1780 & 0.2762 & 0.4037
& 0.5703 & 0.0006
& 0.0831 & 0.2131
& 0.1939 & 0.1761
& 0.1381 & 0.1607
& 0.0361 & 0.9194 \\
& & w/o Txt.
& \graycell 0.7235 & \graycell 0.0020
& \graycell --- & \graycell --- & \graycell 0.3516
& \graycell 0.6728 & \graycell 0.0009
& \graycell 0.0302 & \graycell 0.5306
& \graycell 0.5532 & \graycell 0.1546
& \graycell 0.1747 & \graycell 0.2218
& \graycell 0.0605 & \graycell 0.9194 \\
& & FewShot.
& 0.7185 &  0.0010
& 0.3296 & 0.4023 & 0.5293
& 0.6605 &  0.0000
& 0.1652 & 0.1618
& 0.5575 & 0.2124
& 0.2621 & 0.2913
& 0.0631 & 0.9194 \\

\midrule

\multirow{3}{*}{Gemini 2.5 Flash} & - & Real
& 0.6636 & 0.0000 & 0.2411 & 0.3439 & 0.3916
& 0.7062 & 0.0000 & 0.3265 & 0.3481
& 0.2744 & 0.2507
& 0.1945 & 0.1479
& 0.0418 & 0.6505 \\
& & Real Low R.
& \graycell 0.5596 & \graycell 0.0005
& \graycell 0.0936 & \graycell 0.1485 & \graycell 0.2708
& \graycell 0.5364 & \graycell 0.0000
& \graycell 0.2527 & \graycell 0.2505
& \graycell 0.1581 & \graycell 0.1435
& \graycell 0.1274 & \graycell 0.1065
& \graycell 0.0232 & \graycell 0.5753 \\
& & Synth.
& 0.5622 & 0.0000
& 0.1740 & 0.2909 & 0.3155
& 0.4705 & 0.0000
& 0.1045 & 0.1620
& 0.1766 & 0.1612
& 0.1262 & 0.1321
& 0.0219 & 0.7151 \\
& & w/o Txt.
& \graycell 0.7241 & \graycell 0.0000
& \graycell --- & \graycell --- & \graycell 0.3623
& \graycell 0.7764 & \graycell 0.0000
& \graycell 0.2506 & \graycell 0.0014
& \graycell 0.7044 & \graycell 0.1380
& \graycell 0.2085 & \graycell 0.2539
& \graycell 0.0622 & \graycell 0.5806 \\
& & FewShot.
& 0.7469 &  0.0000
& 0.4956 & 0.4953 & 0.4128
& 0.6910 &  0.0000
& 0.2341 & 0.4517
& 0.7481 & 0.3231
& 0.2071 & 0.2452
& -0.0965 & 0.3226 \\
\midrule

\multirow{3}{*}{Gemini 3.1 Pro-Preview}  & - & Real
& 0.5900 & 0.0003 & 0.2659 & 0.3365 & 0.3523
& 0.6928 & 0.0002 & 0.1710 & 0.2614
& 0.2706 & 0.2490
& 0.1771 & 0.1062
& 0.0432 & 0.6774 \\
& & Real Low R.
& \graycell 0.3992 & \graycell 0.0004
& \graycell 0.0918 & \graycell 0.1434 & \graycell 0.1994
& \graycell 0.4077 & \graycell 0.0001
& \graycell 0.0715 & \graycell 0.0904
& \graycell 0.1101 & \graycell 0.1014
& \graycell 0.0844 & \graycell 0.0721
& \graycell 0.0138 & \graycell 0.9516 \\
& & Synth.
& 0.4805 & 0.0000
& 0.2125 & 0.2444 & 0.2700
& 0.5571 & 0.0001
& 0.1278 & 0.1703
& 0.2108 & 0.1928
& 0.1290 & 0.0876
& 0.0216 & 0.7849 \\
& & w/o Txt.
& \graycell 0.5639 & \graycell 0.0002
& \graycell --- & \graycell --- & \graycell 0.3039
& \graycell 0.5817 & \graycell 0.0000
& \graycell 0.1599 & \graycell 0.2142
& \graycell 0.4041 & \graycell 0.1271
& \graycell 0.0404 & \graycell 0.1609
& \graycell 0.1904 & \graycell 0.7903 \\
& & FewShot.
& 0.6683 &  0.0000
& 0.3113 & 0.2897 & 0.4447
& 0.6363 &  0.0000
& 0.2295 & 0.3472
& 0.4657 & 0.2615
& 0.1656 & 0.1853
& 0.1408 & 0.9140 \\
\midrule

\rowcolor{gray!20}
\multicolumn{18}{c}{Open-Source MLLMs} \\
\midrule

\multirow{3}{*}{LLaMA 4 Maverick} & 400B  & Real
& 0.2527 & 0.0000 & 0.0584 & 0.0821 & 0.1055
& 0.2405 & 0.0000 & 0.0837 & 0.0783
& 0.0672 & 0.0612
& 0.0475 & 0.0434
& 0.0139 & 0.8656 \\
& & Real Low R.
& \graycell 0.1134 & \graycell 0.0002
& \graycell 0.0156 & \graycell 0.0215 & \graycell 0.0401
& \graycell 0.0984 & \graycell 0.0000
& \graycell 0.0311 & \graycell 0.0294
& \graycell 0.0222 & \graycell 0.0204
& \graycell 0.0162 & \graycell 0.0224
& \graycell 0.0045 & \graycell 0.9785 \\
& & Synth.
& 0.2911 & 0.0000
& 0.0667 & 0.0941 & 0.1265
& 0.2868 & 0.0000
& 0.1088 & 0.0962
& 0.0827 & 0.0757
& 0.0575 & 0.0474
& 0.0168 & 0.9301 \\
& & w/o Txt.
& \graycell 0.1956 & \graycell 0.0000
& \graycell --- & \graycell --- & \graycell 0.0350
& \graycell 0.2095 & \graycell 0.0000
& \graycell 0.0123 & \graycell 0.1171
& \graycell 0.1198 & \graycell 0.0185
& \graycell 0.0020 & \graycell 0.0470
& \graycell 0.0052 & \graycell 0.9516 \\
& & FewShot.
& 0.3057 &  0.0006
& 0.1102 & 0.1108 & 0.1679
& 0.3063 &  0.0000
& 0.0557 & 0.0149
& 0.1864 & 0.0380
& 0.0770 & 0.0919
& 0.0163 & 0.2849 \\

\midrule

\multirow{3}{*}{Ministral 14B} & 14B & Real
& 0.3826 & 0.0020 & 0.0867 & 0.1288 & 0.1787
& 0.3794 & 0.0000 & 0.0976 & 0.1219
& 0.1151 & 0.1053
& 0.0825 & 0.1223
& 0.0236 & 0.4409 \\
& & Real Low R.
& \graycell 0.2628 & \graycell 0.0003
& \graycell 0.0223 & \graycell 0.0452 & \graycell 0.0875
& \graycell 0.1696 & \graycell 0.0000
& \graycell 0.0440 & \graycell 0.0529
& \graycell 0.0425 & \graycell 0.0392
& \graycell 0.0314 & \graycell 0.0505
& \graycell 0.0088 & \graycell 0.9785 \\
& & Synth.
& 0.4562 & 0.0000
& 0.0981 & 0.1715 & 0.2464
& 0.3825 & 0.0000
& 0.1244 & 0.1515
& 0.1268 & 0.1152
& 0.0888 & 0.0879
& 0.0288 & 0.5914 \\
& & w/o Txt.
& \graycell 0.4164 & \graycell 0.0011
& \graycell --- & \graycell --- & \graycell 0.0691
& \graycell 0.4697 & \graycell 0.0000
& \graycell 0.2462 & \graycell 0.0303
& \graycell 0.3376 & \graycell 0.0317
& \graycell 0.0068 & \graycell 0.1474
& \graycell 0.0336 & \graycell 0.4839 \\
& & FewShot.
& 0.5054 &  0.0004
& 0.1733 & 0.2229 & 0.3216
& 0.4401 &  0.0000
& 0.1854 & 0.0566
& 0.3707 & 0.0531
& 0.1184 & 0.1407
& 0.0060 & 0.3548 \\

\midrule

\multirow{3}{*}{Qwen3-VL 8B} & 8B & Real
& 0.0868 & 0.0000 & 0.0202 & 0.0354 & 0.0282
& 0.0195 & 0.0000 & 0.0030 & 0.0037
& 0.0043 & 0.0040
& 0.0023 & 0.0065
& -0.0022 & 0.7473 \\
& & Real Low R.
& \graycell 0.0544 & \graycell 0.0000
& \graycell 0.0072 & \graycell 0.0102 & \graycell 0.0109
& \graycell 0.0085 & \graycell 0.0000
& \graycell 0.0024 & \graycell 0.0015
& \graycell 0.0019 & \graycell 0.0017
& \graycell 0.0013 & \graycell 0.0045
& \graycell -0.0006 & \graycell 0.7526 \\
& & Synth.
& 0.0756 & 0.0000
& 0.0166 & 0.0258 & 0.0221
& 0.0020 & 0.0000
& 0.0001 & 0.0002
& 0.0005 & 0.0004
& 0.0002 & 0.0053
& -0.0004 & 1.0000 \\
& & w/o Txt.
& \graycell 0.0577 & \graycell 0.0002
& \graycell --- & \graycell --- & \graycell 0.0032
& \graycell 0.0083 & \graycell 0.0000
& \graycell 0.0000 & \graycell 0.0000
& \graycell 0.0042 & \graycell 0.0003
& \graycell 0.0000 & \graycell 0.0046
& \graycell 0.0000 & \graycell 0.8118 \\
& & FewShot.
& 0.0847 &  0.0000
& 0.0324 & 0.0325 & 0.0378
& 0.0275 &  0.0000
& 0.0052 & 0.0036
& 0.0156 & 0.0020
& 0.0036 & 0.0079
& 0.0000 & 0.3871 \\



\midrule

\multirow{3}{*}{Qwen3.5-122B-A10B} & 122B & Real
& 0.5584 & 0.0008 & 0.2442 & 0.2826 & 0.3464
& 0.5164 & 0.0000 & 0.1957 & 0.1321
& 0.4237 & 0.1230
& 0.1179 & 0.1452
& 0.0067 & 0.9140 \\
& & Real Low R.
& \graycell 0.3190 & \graycell 0.0003
& \graycell 0.0715 & \graycell 0.0920 & \graycell 0.1706
& \graycell 0.2631 & \graycell 0.0000
& \graycell 0.0693 & \graycell 0.0460
& \graycell 0.1831 & \graycell 0.0339
& \graycell 0.0871 & \graycell 0.1018
& \graycell 0.0361 & \graycell 0.9301 \\
& & Synth.
& 0.5137 & 0.0000
& 0.2440 & 0.2577 & 0.3215
& 0.5406 & 0.0000
& 0.1661 & 0.1149
& 0.4301 & 0.1471
& 0.2513 & 0.2711
& 0.0000 & 0.7096 \\
& & w/o Txt.
& \graycell 0.3649 & \graycell 0.0023
& \graycell --- & \graycell --- & \graycell 0.1719
& \graycell 0.3459 & \graycell 0.0002
& \graycell 0.0789 & \graycell 0.0305
& \graycell 0.2499 & \graycell 0.0630
& \graycell 0.1844 & \graycell 0.1947
& \graycell 0.0000 & \graycell 0.6828 \\
& & FewShot.
& 0.6975 & 0.0008
& 0.3564 & 0.4373 & 0.5328
& 0.6506 & 0.0003
& 0.3341 & 0.3198
& 0.5480 & 0.2068
& 0.2837 & 0.3243
& 0.0140 & 0.9086 \\

\midrule

\multirow{3}{*}{Classical Vision Pipeline} & - & Real
& 0.4294 & 0.8812 & 0.5078 & 0.2487 & 0.7875
& 0.1502 & 0.9174 & 0.6245 & 0.4100
& 0.5871 & 0.6613
& 0.1060 & 0.0311
& 0.0442 & 1.0000 \\
& & Real Low R.
& \graycell 0.9051 & \graycell 0.9433
& \graycell 0.5055 & \graycell 0.2983 & \graycell 0.8235
& \graycell 0.0000 & \graycell 0.8503
& \graycell 0.7258 & \graycell 0.5288
& \graycell 0.6156 & \graycell 0.4951
& \graycell --- & \graycell ---
& \graycell --- & \graycell 1.0000 \\
& & Synth.
& 0.4240 & 0.9518
& 0.6087 & 0.5977 & 0.8319
& 0.1960 & 0.9047
& 0.8170 & 0.7347
& 0.7045 & 0.7031
& 0.2280 & 0.0397
& 0.1263 & 1.0000 \\
& & Synth. w/o Txt.
& \graycell --- & \graycell 0.9199
& \graycell --- & \graycell --- & \graycell 0.8317
& \graycell --- & \graycell 0.8554
& \graycell 0.9812 & \graycell 0.8536
& \graycell 0.7081 & \graycell 0.8065
& \graycell --- & \graycell ---
& \graycell --- & \graycell 1.0000 \\

\midrule

\multirow{3}{*}{Human Engineer} & - & Real
& 0.9667 & 0.9589 & 0.9842 & 0.9759 & 0.9547
& 0.9534 & 0.8793 & 0.9987 & 1.0
& 0.9867 & 0.9353
& 0.8696 & 0.9188
& 0.8876 & 1.0000 \\
& & Synth.
& \graycell 0.9649 & \graycell 0.9678
& \graycell 0.9851 & \graycell 0.9748 & \graycell 0.9521
& \graycell 0.9516 & \graycell 0.9732
& \graycell 0.9989 & \graycell 0.9992
& \graycell 0.9893 & \graycell 0.9867
& \graycell 0.8473 & \graycell 0.9136
& \graycell 0.8846 & \graycell 1.0000 \\

\bottomrule
\end{tabular}%
}
\endgroup
\end{table*}

%% file: eval_task3_3.tex
\begin{table*}[t]
\centering
\caption{\textbf{Evaluation results of agentic evaluation.} \textit{w/o GT BBox} denotes full-image search, where LMMs must first localize the target before performing visual search; \textit{w/ GT BBox} denotes detector-assisted search, where an initial bounding box is provided and LMMs only expand it without re-localizing the target. Both input images are the generated schematic in Fig.~\ref{fig:synth_data}(d). $\Delta$ denotes the relative improvement (\%) of \textit{w/ GT BBox} vs. \textit{w/o GT BBox}; \textit{*} denotes metrics computed only on instances with visible text.}
\label{tab:combined_results}

\scriptsize
\setlength{\tabcolsep}{3pt}
\renewcommand{\arraystretch}{1.1}

\resizebox{\textwidth}{!}{
\begin{tabular}{l l l ccccc ccccc c}
\toprule

Model & Size & Setting
& \multicolumn{5}{c|}{Symbol}
& \multicolumn{4}{c|}{Pin}
& \multicolumn{1}{c}{Net} \\

& &
& Name & Name* & Value & Value* & Type
& PinName & PinName* & PadName & PadName*
& NetName \\

\midrule

\rowcolor{gray!20}
\multicolumn{13}{c}{Commercial chatbot systems} \\
\midrule

\multirow{3}{*}{GPT-5.2}
& - & w/o GT BBox
& \second{0.6948} & 0.6457 & 0.6610 & 0.6507 & 0.6886
& \best{0.7892} & \second{0.7621} & 0.2654 & 0.1298
& -- \\
& & w/ GT BBox
& \graycell 0.8565 & \graycell 0.8482 & \graycell 0.8088 & \graycell 0.8252 & \graycell 0.8458
& \graycell \best{0.9056} & \graycell \second{0.8567} & \graycell \best{0.8426} & \graycell \second{0.9473}
& \graycell \second{0.6926} \\
& & $\Delta$
& \pos{23.28} & \pos{31.36} & \pos{22.36} & \pos{26.82} & \pos{22.83}
& \pos{14.76} & \pos{12.41} & \pos{217.57} & \pos{629.43}
& -- \\

\midrule

\multirow{3}{*}{GPT-5 Mini}
& - & w/o GT BBox
& 0.5539 & 0.5835 & 0.4983 & 0.4797 & 0.5292
& 0.6285 & 0.5545 & 0.1311 & 0.1138
& -- \\
& & w/ GT BBox
& \graycell \second{0.8932} & \graycell \second{0.8739} & \graycell \second{0.8437} & \graycell \second{0.8799} & \graycell \second{0.8717}
& \graycell 0.8694 & \graycell 0.7615 & \graycell 0.7839 & \graycell 0.8196
& \graycell \second{0.6517} \\
& & $\Delta$
& \pos{61.26} & \pos{45.36} & \pos{69.31} & \pos{72.01} & \pos{64.72}
& \pos{38.33} & \pos{37.34} & \pos{497.94} & \pos{620.21}
& -- \\

\midrule

\multirow{3}{*}{Claude-Sonnet-4.6}
& - & w/o GT BBox
& 0.5070 & \second{0.7414} & \second{0.6710} & \second{0.7368} & \second{0.7452}
& 0.4982 & 0.5944 & 0.4022 & \second{0.6210}
& -- \\
& & w/ GT BBox
& \graycell 0.7955 & \graycell 0.7210 & \graycell 0.7430 & \graycell 0.7589 & \graycell 0.7202
& \graycell \second{0.8984} & \graycell \best{0.8864} & \graycell 0.6400 & \graycell \best{0.9526}
& \graycell 0.6411 \\
& & $\Delta$
& \pos{56.91} & \pos{0.71} & \pos{10.73} & \pos{3.00} & \negv{3.36}
& \pos{80.33} & \pos{49.11} & \pos{59.15} & \pos{53.38}
& -- \\

\midrule

\multirow{3}{*}{Gemini-2.5-Flash-Lite}
& - & w/o GT BBox
& 0.2148 & 0.3991 & 0.2280 & 0.2129 & 0.2482
& 0.1052 & 0.1504 & 0.0998 & 0.0708
& -- \\
& & w/ GT BBox
& \graycell 0.5123 & \graycell 0.7467 & \graycell 0.5181 & \graycell 0.4776 & \graycell 0.6156
& \graycell 0.8473 & \graycell 0.7870 & \graycell 0.7184 & \graycell 0.7220
& \graycell 0.3873 \\
& & $\Delta$
& \pos{138.49} & \pos{87.06} & \pos{127.24} & \pos{124.33} & \pos{148.02}
& \pos{705.42} & \pos{423.01} & \pos{619.64} & \pos{920.34}
& -- \\

\midrule

\multirow{3}{*}{Gemini-3.1-Pro-Preview}
& - & w/o GT BBox
& \best{0.8093} & \best{0.8726} & \best{0.7833} & \best{0.8151} & \best{0.7870}
& \second{0.6943} & \best{0.8211} & \best{0.5824} & \best{0.8927}
& -- \\
& & w/ GT BBox
& \graycell \best{0.8958} & \graycell \best{0.8983} & \graycell \best{0.8591} & \graycell \best{0.9013} & \graycell \best{0.8764}
& \graycell 0.8382 & \graycell 0.8140 & \graycell \second{0.7841} & \graycell 0.8545
& \graycell \best{0.8911} \\
& & $\Delta$
& \pos{10.69} & \pos{2.95} & \pos{9.68} & \pos{10.58} & \pos{11.36}
& \pos{20.72} & \negv{0.86} & \pos{34.62} & \negv{4.28}
& -- \\

\midrule

\rowcolor{gray!20}
\multicolumn{13}{c}{Open-source MLLMs} \\
\midrule

\multirow{3}{*}{LLaMA-4-Maverick-400B}
& 400B & w/o GT BBox
& 0.1270 & 0.1699 & 0.1389 & 0.0876 & 0.2034
& 0.3346 & 0.0455 & 0.1780 & 0.0273
& -- \\
& & w/ GT BBox
& \graycell 0.7934 & \graycell 0.7229 & \graycell 0.7421 & \graycell 0.7579 & \graycell 0.7196
& \graycell 0.7586 & \graycell 0.4964 & \graycell 0.5276 & \graycell 0.5323
& \graycell 0.3758 \\
& & $\Delta$
& \pos{524.72} & \pos{325.37} & \pos{434.41} & \pos{765.41} & \pos{253.77}
& \pos{126.70} & \pos{991.21} & \pos{196.40} & \pos{1849.45}
& -- \\

\midrule

\multirow{3}{*}{Ministral-14B}
& 14B & w/o GT BBox
& 0.0990 & 0.0784 & 0.1324 & 0.0421 & 0.1031
& 0.1725 & 0.0380 & 0.0524 & 0.0051
& -- \\
& & w/ GT BBox
& \graycell 0.4179 & \graycell 0.7626 & \graycell 0.3882 & \graycell 0.3622 & \graycell 0.4399
& \graycell 0.3960 & \graycell 0.2742 & \graycell 0.3457 & \graycell 0.1301
& \graycell 0.1113 \\
& & $\Delta$
& \pos{322.12} & \pos{872.70} & \pos{193.25} & \pos{760.57} & \pos{326.61}
& \pos{129.57} & \pos{621.58} & \pos{559.92} & \pos{2450.98}
& -- \\

\midrule

\multirow{3}{*}{Qwen3-8B-Instruct}
& 8B & w/o GT BBox
& 0.2337 & 0.0368 & 0.1126 & 0.0280 & 0.0361
& 0.6653 & 0.0106 & \second{0.5544} & 0.0013
& -- \\
& & w/ GT BBox
& \graycell 0.5917 & \graycell 0.7057 & \graycell 0.5872 & \graycell 0.5725 & \graycell 0.6067
& \graycell 0.7418 & \graycell 0.5723 & \graycell 0.6207 & \graycell 0.4153
& \graycell 0.3021 \\
& & $\Delta$
& \pos{153.20} & \pos{2039.67} & \pos{421.40} & \pos{1948.21} & \pos{1581.99}
& \pos{11.50} & \pos{5299.06} & \pos{11.96} & \pos{31792.31}
& -- \\

\midrule

\multirow{3}{*}{Qwen3-VL-235B-A22B}
& 235B & w/o GT BBox
& 0.1336 & 0.2315 & 0.1570 & 0.1187 & 0.2150
& 0.2099 & 0.1027 & 0.0909 & 0.0130
& -- \\
& & w/ GT BBox
& \graycell 0.5753 & \graycell 0.7871 & \graycell 0.5792 & \graycell 0.5809 & \graycell 0.6566
& \graycell 0.7282 & \graycell 0.8283 & \graycell 0.7233 & \graycell 0.8380
& \graycell 0.4532 \\
& & $\Delta$
& \pos{330.64} & \pos{277.44} & \pos{268.92} & \pos{640.69} & \pos{205.39}
& \pos{246.88} & \pos{706.40} & \pos{695.99} & \pos{6346.15}
& -- \\

\bottomrule
\end{tabular}
}
\end{table*}